\documentclass{article}
\usepackage{spconf,amsmath,graphicx}

\usepackage{amsfonts}       
\usepackage{nicefrac}       
\usepackage{microtype}      
\usepackage{color}
\usepackage{multirow}
\usepackage{booktabs}
\RequirePackage{lineno}

\title{SA-Net: Shuffle Attention for Deep Convolutional Neural Networks}
\name{Qing-Long Zhang, Yu-Bin Yang\sthanks{Funded by the Natural Science Foundation of China (No. 61673204).}}
\address{State Key Laboratory for Novel Software Technology at Nanjing University}
\begin{document}
%
\maketitle
\begin{abstract}
Attention mechanisms, which enable a neural network to accurately focus on all the relevant elements of the input, have become an essential component to improve the performance of deep neural networks. There are mainly two attention mechanisms widely used in computer vision studies, \textit{spatial attention} and \textit{channel attention}, which aim to capture the pixel-level pairwise relationship and channel dependency, respectively. Although fusing them together may achieve better performance than their individual implementations, it will inevitably increase the computational overhead. In this paper, we propose an efficient Shuffle Attention (SA) module to address this issue, which adopts Shuffle Units to combine two types of attention mechanisms effectively. Specifically, SA first groups channel dimensions into multiple sub-features before processing them in parallel. Then, for each sub-feature, SA utilizes a Shuffle Unit to depict feature dependencies in both spatial and channel dimensions. After that, all sub-features are aggregated and a ``channel shuffle" operator is adopted to enable information communication between different sub-features.
The proposed SA module is efficient yet effective, e.g., the parameters and computations of SA against the backbone ResNet50 are 300 vs. 25.56M and 2.76e-3 GFLOPs vs. 4.12 GFLOPs, respectively, and the performance boost is more than 1.34\% in terms of Top-1 accuracy.
Extensive experimental results on common-used benchmarks, including ImageNet-1k for classification, MS COCO for object detection, and instance segmentation, demonstrate that the proposed SA outperforms the current SOTA methods significantly by achieving higher accuracy while having lower model complexity.  The code and models are available at https://github.com/wofmanaf/SA-Net.

\end{abstract}

\section{Introduction}
\label{sec:intro}

\begin{figure}[h]
	\centering
	\includegraphics[width=1.0\linewidth]{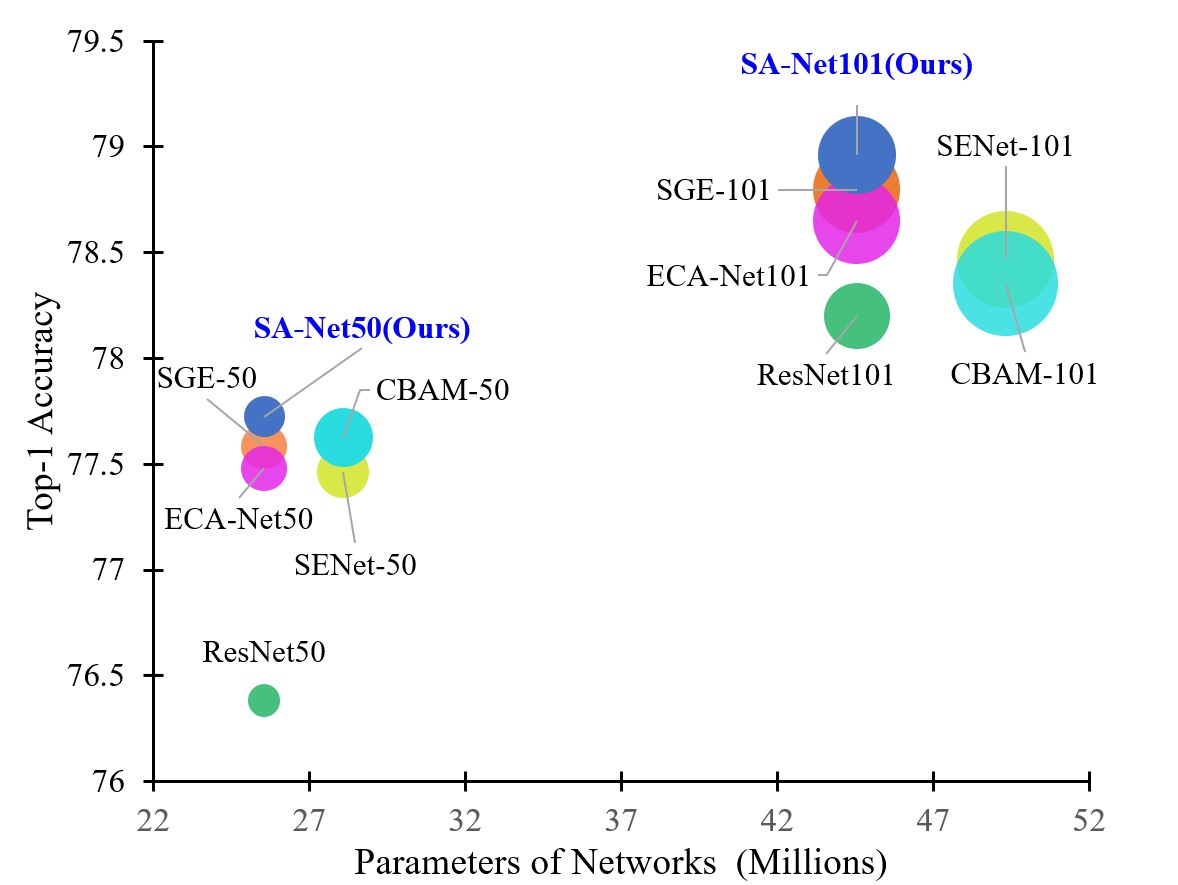}
	\caption{Comparisons of recently SOTA attention models on ImageNet-1k, including SENet, CBAM, ECA-Net, SGE-Net, and SA-Net, using ResNets as backbones, in terms of accuracy, network parameters, and GFLOPs. The size of circles indicates the GFLOPs. Clearly, the proposed SA-Net achieves higher accuracy while having less model complexity.}
	\label{fig:figure1}
\end{figure}

Attention mechanisms have been attracting increasing attention in research communities since it helps to improve the representation of interests, i.e., focusing on essential features while suppressing unnecessary ones \cite{DBLP:journals/corr/abs-1904-11492,DBLP:conf/cvpr/LiW0019,DBLP:conf/cvpr/FuLT0BFL19,DBLP:journals/corr/abs-1809-00916}. 
Recent studies show that correctly incorporating attention mechanisms into convolution blocks can significantly improve the performance of a broad range of computer vision tasks, e.g., image classification, object detection, and instance segmentation.

There are mainly two types of attention mechanisms most commonly used in computer vision: \textit{channel attention} and \textit{spatial attention}, both of which strengthen the original features by aggregating the same feature from all the positions with different aggregation strategies, transformations, and strengthening functions \cite{DBLP:journals/corr/abs-1903-10829,DBLP:conf/eccv/ZhaoZLSLLJ18,DBLP:journals/corr/abs-1907-13426,DBLP:journals/corr/abs-1908-07678,DBLP:journals/corr/abs-1811-11721}.
Based on these observations, some studies, including GCNet \cite{DBLP:journals/corr/abs-1904-11492} and CBAM \cite{DBLP:conf/eccv/WooPLK18} integrated both \textit{spatial attention} and \textit{channel attention} into one module and achieving significant improvement~\cite{DBLP:conf/eccv/WooPLK18,DBLP:journals/corr/abs-1809-00916}. However, they generally suffered from either converging difficulty or heavy computation burdens. Other researches managed to simplify the structure of channel or spatial attention \cite{DBLP:journals/corr/abs-1904-11492,DBLP:conf/cvpr/WangWZLZH20}.
For example, ECA-Net~\cite{DBLP:conf/cvpr/WangWZLZH20} simplifies the process of computing channel weights in SE block by using a 1-D convolution. SGE~\cite{DBLP:journals/corr/abs-1905-09646} groups the dimension of channels into multiple sub-features to represent different semantics, and applies a spatial mechanism to each feature group by scaling the feature vectors over all locations with an attention mask.
However, they did not take full advantage of the correlation between spatial and channel attention, making them less efficient.
``Can one fuse different attention modules in a lighter but more efficient way?"

To answer this question, we first revisit the unit of ShuffleNet v2~\cite{DBLP:conf/eccv/MaZZS18}, which can efficiently construct a multi-branch structure and process different branches in parallel. Specifically, at the beginning of each unit, the input of $c$ feature channels are split into two branches with $c-c'$ and $c'$ channels. Afterwards, several convolution layers are adopted to capture a higher-level representation of the input. After these convolutions, the two branches are concatenated to make the number of channels as same as the number of input. At last, the ``channel shuffle" operator (defined in \cite{DBLP:conf/cvpr/ZhangZLS18}) is adopted to enable information communication between the two branches.
In addition, to increase calculation speed, SGE~\cite{DBLP:journals/corr/abs-1905-09646} introduces a grouping strategy, which divides the input feature map into groups along the channel dimension. Then all sub-features can be enhanced parallelly.

Based on these above observations, this paper proposes a lighter but more efficient Shuffle Attention (SA) module for deep Convolutional Neural Networks(CNNs), which groups the dimensions of channel into sub-features.
For each sub-feature, SA adopts the Shuffle Unit to construct \textit{channel attention} and \textit{spatial attention} simultaneously. For each attention module, this paper designs an attention mask over all the positions to suppress the possible noises and highlight the correct semantic feature regions as well. Experimental results on ImageNet-1k (which are shown in Figure~\ref{fig:figure1}) have shown that the proposed simple but effective module containing fewer parameters has achieved higher accuracy than the current state-of-the-art methods.

The key contributions in this paper are summarized as follows:
1) we introduce a lightweight yet effective attention module, SA, for deep CNNs, which groups channel dimensions into multiple sub-features, and then utilizes a Shuffle Unit to integrate the complementary channel and spatial attention module for each sub-feature.
2) extensive experimental results on ImageNet-1k and MS COCO demonstrate that the proposed SA has lower model complexity than the state-of-the-art attention approaches while achieving outstanding performance.

\section{Related Work}
{\bf Multi-branch architectures.} Multi-branch architectures of CNNs have evolved for years and are becoming more accurate and faster. The principle behind multi-branch architectures is ``split-transform-merge", which eases the difficulty of training networks with hundreds of layers. The InceptionNet series~\cite{DBLP:conf/cvpr/SzegedyVISW16,DBLP:conf/aaai/SzegedyIVA17} are successful multi-branch architectures of which each branch is carefully configured with customized kernel filters, in order to aggregate more informative and multifarious features. ResNets~\cite{DBLP:conf/cvpr/HeZRS16} can also be viewed as two-branch networks, in which one branch is the identity mapping. SKNets~\cite{DBLP:conf/cvpr/LiW0019} and ShuffleNet families~\cite{DBLP:conf/eccv/MaZZS18} both followed the idea of InceptionNets with various filters for multiple branches while differing in at least two important aspects. SKNets utilized an adaptive selection mechanism to realize adaptive receptive field size of neurons. ShuffleNets further merged ``channel split" and ``channel shuffle" operators into a single element-wise operation to make a trade-off between speed and accuracy.

{\bf Grouped Features.}
Learning features into groups dates back to AlexNet~\cite{DBLP:conf/nips/KrizhevskySH12}, whose motivation is distributing the model over more GPU resources. Deep Roots examined AlexNet and pointed out that convolution groups can learn better feature representations. The MobileNets~\cite{DBLP:conf/cvpr/SandlerHZZC18,DBLP:conf/iccv/HowardPALSCWCTC19} and ShuffleNets~\cite{DBLP:conf/eccv/MaZZS18} treated each channel as a group, and modeled the spatial relationships within these groups. CapsuleNets~\cite{DBLP:conf/nips/SabourFH17,DBLP:conf/iclr/HintonSF18} modeled each grouped neuron as a capsule, in which the neuron activity in the active capsule represented various attributes of a particular entity in the image. SGE~\cite{DBLP:journals/corr/abs-1905-09646} developed CapsuleNets and divided the dimensions of channel into multiple sub-features to learn different semantics.

{\bf Attention mechanisms.}
The significance of attention has been studied extensively in the previous literature. It biases the allocation of the most informative feature expressions while suppressing the less useful ones. The self-attention method calculates the context in one position as a weighted sum of all the positions in an image. SE~\cite{DBLP:conf/cvpr/HuSS18} modeled channel-wise relationships using two FC layers. ECA-Net~\cite{DBLP:conf/cvpr/WangWZLZH20} adopted a 1-D convolution filter to generate channel weights and significantly reduced the model complexity of SE. Wang et al.~\cite{DBLP:conf/cvpr/0004GGH18} proposed the non-local(NL) module to generate a considerable attention map by calculating the correlation matrix between each spatial point in the feature map.
CBAM~\cite{DBLP:conf/eccv/WooPLK18}, GCNet~\cite{DBLP:journals/corr/abs-1904-11492}, and SGE~\cite{DBLP:journals/corr/abs-1905-09646} combined the \textit{spatial attention} and \textit{channel attention} serially, while DANet~\cite{DBLP:conf/cvpr/FuLT0BFL19} adaptively integrated local features with their global dependencies by summing the two attention modules from different branches.

\section{Shuffle Attention}
\begin{figure*}[htb]
	\centering
	\includegraphics[width=1.0\linewidth]{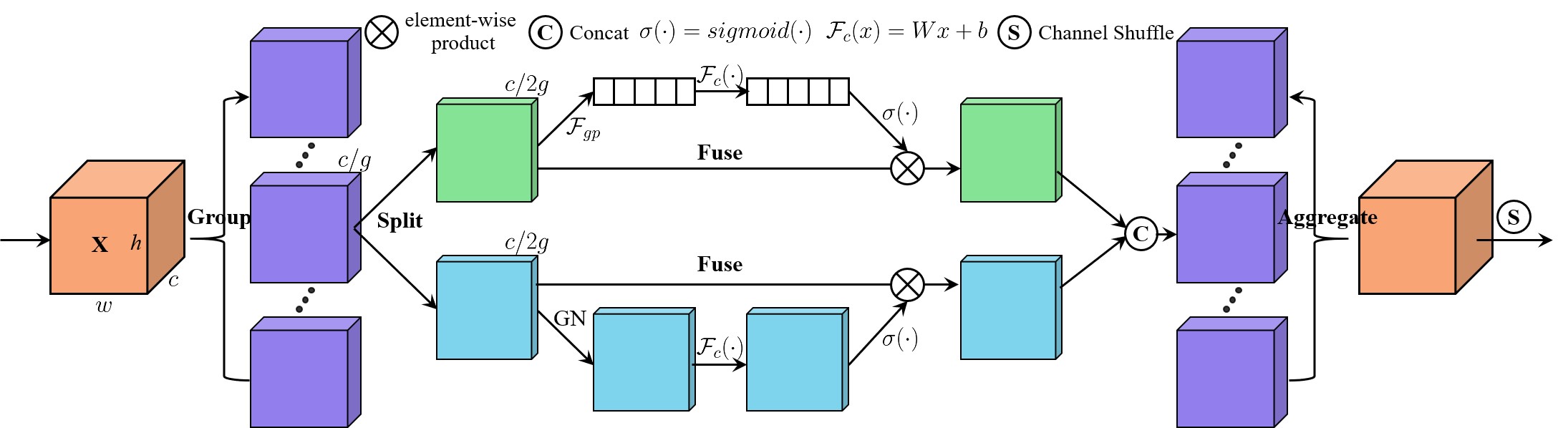}
	\caption{An overview of the proposed SA module. It adopts ``channel split" to process the sub-features of each group in parallel. For channel attention branch, using GAP to generate channel-wise statistics, then use a pair of parameters to scale and shift the channel vector. For spatial attention branch, adopting group norm to generate spatial-wise statistics, then a compact feature is created similar to the channel branch. The two branches are then concatenated. After that, all sub-features are aggregated and finally we utilize a ``channel shuffle" operator to enable information communication between different sub-features.}
	\label{fig:figure2}
\end{figure*}

In this section, we firstly introduce the process of constructing the SA module, which divides the input feature map into groups, and uses Shuffle Unit to integrate the \textit{channel attention} and \textit{spatial attention} into one block for each group. After that, all sub-features are aggregated and a ``channel shuffle" operator is utilized to enable information communication between different sub-features. Then, we show how to adopt SA for deep CNNs. Finally, we visualize the effect and validate the reliability of the proposed SA. The overall architecture of SA module is illustrated in Figure~\ref{fig:figure2}.

{\bf Feature Grouping.} For a given feature map $X \in \mathbb{R}^{C\times H \times W}$, where $C$, $H$, $W$ indicate the channel number, spatial height, and width, respectively, SA first divides $X$ into $G$ groups along the channel dimension, i.e., $X=[X_1, \cdots, X_G], X_k \in \mathbb{R}^{C/G\times H \times W}$, in which each sub-feature $X_k$ gradually captures a specific semantic response in the training process. Then, we generate the corresponding importance coefficient for each sub-feature through an attention module. Specifically, at the beginning of each attention unit, the input of $X_k$ is split into two branches along the channels dimension, i.e., $X_{k1}$,  $X_{k2} \in \mathbb{R}^{C/2G\times H \times W}$. As illustrated in Figure~\ref{fig:figure2}, one branch is adopted to produce a channel attention map by exploiting the inter-relationship of channels, while the other branch is used to generate a spatial attention map by utilizing the inter-spatial relationship of features, so that the model can focus on ``what" and ``where" is meaningful.

{\bf Channel Attention.} An option to fully capture channel-wise dependencies is utilizing the SE block proposed in \cite{DBLP:conf/cvpr/HuSS18}. However, it will bring too many parameters, which is not good for designing a more lightweight attention module in terms of a trade-off between speed and accuracy. Also, it is not suitable to generate channel weights by performing a faster 1-D convolution of size $k$ like ECA~\cite{DBLP:conf/cvpr/WangWZLZH20} because $k$ tends to be larger. To improve, we provide an alternative, which firstly embeds the global information by simply using global averaging pooling (GAP) to generate channel-wise statistics as $s\in \mathbb{R}^{C/2G \times 1 \times 1}$, which can be calculated by shrinking $X_{k1}$ through spatial dimension $H \times W$:
\begin{equation}
	s = \mathcal{F}_{gp}(X_{k1}) = \frac{1}{H\times W}\sum_{i=1}^{H}\sum_{j=1}^{W}X_{k1}(i,j)
\end{equation}

Furthermore, a compact feature is created to enable guidance for precise and adaptive selection. This is achieved by a simple gating mechanism with sigmoid activation. Then, the final output of channel attention can be obtained by
\begin{equation}
	X'_{k1} = \sigma (\mathcal{F}_{c}(s)) \cdot X_{k1} = \sigma(W_1 s + b_1)\cdot X_{k1}
\end{equation}
where $W_1\in \mathbb{R}^{C/2G \times 1 \times 1}$ and $b_1 \in \mathbb{R}^{C/2G \times 1 \times 1}$ are parameters used to scale and shift $s$.

\begin{figure}[h]
	\centering
	\includegraphics[width=0.98\linewidth]{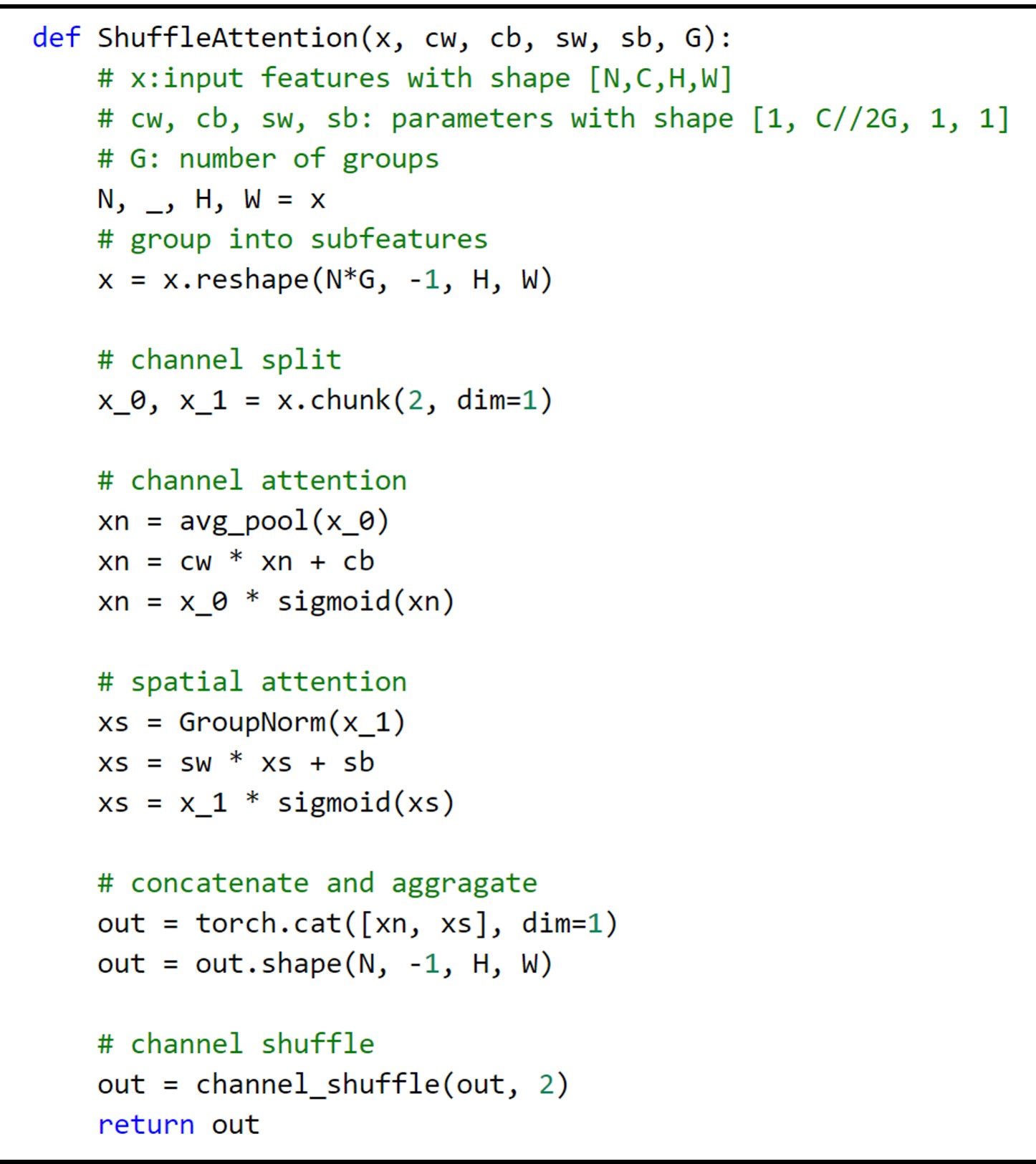}
	\caption{PyTorch code of the proposed SA module}
	\label{fig:code}
\end{figure}

\begin{figure*}[htb]
	\centering
	\includegraphics[width=0.98\linewidth]{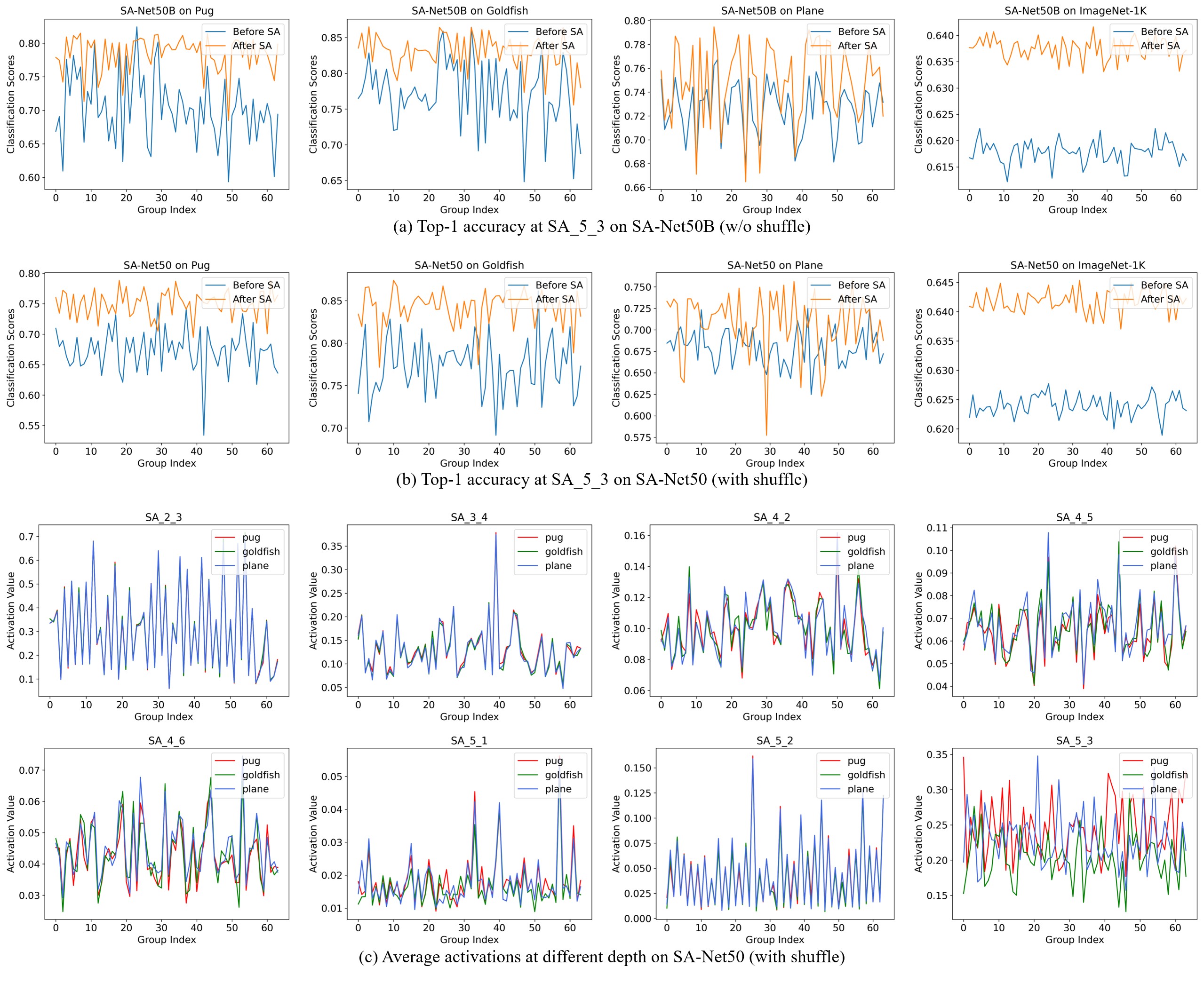}
	\caption{Validation on the effectiveness of SA.}
	\label{fig:figure4}
\end{figure*}

{\bf Spatial Attention.}
Different from channel attention, spatial attention focuses on ``where" is an informative part, which is complementary to channel attention. First, we use Group Norm (GN)\cite{DBLP:conf/eccv/WuH18} over $X_{k2}$ to obtain spatial-wise statistics.
Then, $\mathcal{F}_{c}(\cdot)$ is adopted to enhance the representation of $\hat{X}_{k2}$. The final output of spatial attention is obtained by
\begin{equation}
	X'_{k2} = \sigma(W_2 \cdot GN(X_{k2}) + b_2) \cdot X_{k2}
\end{equation}
where $W_2$ and $b_2$ are parameters with shape $\mathbb{R}^{C/2G \times 1 \times 1}$.

Then the two branches are concatenated to make the number of channels as the same as the number of input, i.e. $X'_k=[X'_{k1},X'_{k2}] \in \mathbb{R}^{C/G\times H \times W}$.

{\bf Aggregation.} After that, all the sub-features are aggregated. And finally, similar to ShuffleNet v2 \cite{DBLP:conf/eccv/MaZZS18}, we adopt a "channel shuffle" operator to enable cross-group information flow along the channel dimension.
The final output of SA module is the same size of $X$, making SA quite easy to be integrated with modern architectures.

Note that $W_1, b_1, W_2, b_2$ and Group Norm hyper-parameters are only parameters introduced in the proposed SA. In a single SA module, the number of channels in each branch is $C/2G$. Therefore, the total parameters are $3C/G$ (typically $G$ is 32 or 64), which is trivial compared with the millions of parameters of the entire network, making SA quite lightweight.
The overall architecture of SA module is illustrated in Figure~\ref{fig:figure2}.

{\bf Implementation.} SA can be easily implemented by a few lines of code in PyTorch and TensorFlow where automatic differentiation is supported. Figure~\ref{fig:code} shows the code based on PyTorch.

{\bf SA-Net for Deep CNNs.} For adopting SA into deep CNNs, we exploit exactly the same configuration with SENet~\cite{DBLP:conf/cvpr/HuSS18} and just replace SE block with SA module. The generated networks are named as SA-Net.

\begin{figure*}[h]
	\centering
	\includegraphics[width=0.95\linewidth]{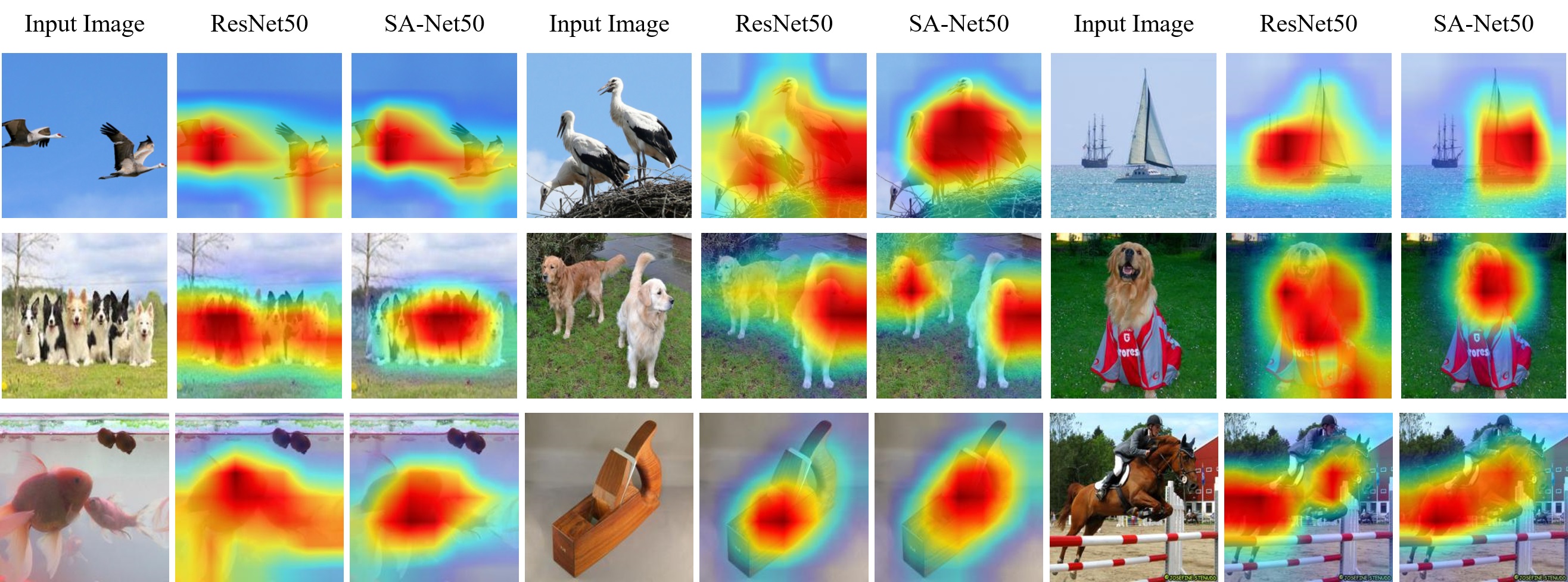}
	\caption{Sample visualization on ImageNet-1k val split generated by GradCAM. All target layer selected is ``layer4.2".}
	\label{fig:figure7}
\end{figure*}

\subsection{Visualization and Interpretation}
In order to verify whether SA can improve semantic feature representation by utilizing feature grouping and channel shuffle, we first train SA-Net50B (without ``channel shuffle") and SA-Net50 (with ``channel shuffle") on the ImageNet-1k training set. Assume $I$ is the original input, we calculate Top-1 accuracy of $I\times X_k$ in each group at SA\_5\_3 (i.e., the last bottleneck according to the following scheme: SA\_stageID\_blockID.) before and after using of the SA module. We use the accuracy scores as indicator and plot the distribution for different classes (``pug'',``goldfish" and ``plane") in the ImageNet-1k validation set in Figure~\ref{fig:figure4}(a, b). For comparison, we also plot the distribution of classification scores across all the 1000 classes.

As shown in Figure~\ref{fig:figure4} (a, b), the Top-1 accuracy statistically increases after SA, which means feature grouping can significantly enhance the semantic representations of feature maps. In addition, the average score in each group gains ($\approx 0.4\%$) with ``channel shuffle", which demonstrates the effectiveness of ``channel shuffle".

To fully validate the effectiveness of SA, we plot the distribution of average activations (the mean value of the channel-wise feature maps in each group, similar to SE) across three classes (``pug", ``goldfish", and ``plane") at different depths in SA-Net50 (with shuffle). The results are shown in Figure~\ref{fig:figure4}(c). We make some observations about the role of SA module:
(1) the distribution across different classes is very similar to each other at the earlier layers(e.g., SA\_2\_3 and SA\_3\_4),  which suggests that the importance of feature groups is likely to be shared by different classes in the early stages;
(2) at greater depths, the activation of each group becomes much more class-specific as different classes exhibit different performance to the discriminative value of features(e.g., SA\_4\_6 and SA\_5\_3);
(3) SA\_5\_2 exhibits a similar pattern over different classes, which means SA\_5\_2 is less important than other blocks in providing recalibration to the network.

In order to validate the effectiveness of SA more intuitively, we sample 9 images from ImageNet-1k val split. We use GradCAM~\cite{DBLP:conf/iccv/SelvarajuCDVPB17} to visualize their heatmaps at SA\_5\_3 on SA-Net50. For comparison, we also draw their heatmaps of ResNet50 at ``layer4.2". As shown in Figure~\ref{fig:figure7}, our proposed SA module allows the classification model to focus on more relevant regions with more object details, which means the SA module can effectively improve the classification accuracy.

Therefore, the proposed SA module is validated to indeed enhance the representation power of networks.

\section{Experiments}

{\bf Experiment Setup.} All experiments are conducted with  exactly the same data augmentation and hyper-parameters settings in \cite{DBLP:conf/cvpr/HeZRS16} and \cite{DBLP:conf/cvpr/HuSS18}.
Specifically, the input images are randomly cropped to $224 \times 224$ with random horizontal flipping. The numbers of groups $G$ in SA-Net and SGE-Net are both set 64. The initialization of parameters in $\mathcal{F}_c(\cdot)$ are set to 0 for weights ($W_1$ and $W_2$) and 1 for biases ($b_1$ and $b_2$) to obtain better results. All the architectures are trained from scratch by SGD with weight decay 1e-4, momentum 0.9, and mini-batch size 256 (using 8 GPUs with 32 images per GPU) for 100 epochs, starting from the initial learning rate 0.1 (with a linear warm-up~\cite{DBLP:journals/corr/GoyalDGNWKTJH17} of 5 epochs) and decreasing it by a factor of 10 every 30 epochs. For the testing on the validation set, the shorter side of an input image is first resized to 256, and a center crop of $224 \times 224$ is used for evaluation. 

\begin{table*}[htbp]
	\caption{Comparisons of different attention methods on ImageNet-1k in terms of network parameters (Param.), giga floating point operations per second (GFLOPs), and Top-1/Top-5 accuracy (in \%). The best records and the improvements are marked in \textbf{bold} and $\uparrow$, respectively.}
	\label{tab:tab1}
	\begin{center}
		\setlength{\tabcolsep}{4.5mm}{
			\begin{tabular}{c|c|c|c|c|c}
				\toprule[1.2pt]
				Attention Methods & Backbones & Param. & GFLOPs & Top-1 Acc (\%) & Top-5 Acc (\%) \\
				\midrule[1.2pt]
				ResNet \cite{DBLP:conf/cvpr/HeZRS16} & \multirow{6}[0]{*}{ResNet-50} & 25.557M & 4.122 & 76.384 & 92.908 \\
				SENet \cite{DBLP:conf/cvpr/HuSS18} & & 28.088M & 4.130  & 77.462 & 93.696 \\
				CBAM \cite{DBLP:conf/eccv/WooPLK18} & & 28.090M & 4.139 & 77.626 & 93.660 \\
				SGE-Net \cite{DBLP:journals/corr/abs-1905-09646}  & & 25.559M & 4.127 & 77.584 & 93.664 \\
				ECA-Net \cite{wang2020eca} &  & \textbf{25.557M} & 4.127 & 77.480 & 93.680 \\
				\textbf{SA-Net (Ours)}  &       & \textbf{25.557M} & \textbf{4.125} & \textbf{77.724 (\textcolor[rgb]{ 0,  0,  1}{$\uparrow 1.34$})} & \textbf{93.798 (\textcolor[rgb]{ 0,  0,  1}{$\uparrow 0.89$})} \\
				\midrule[1.2pt]
				ResNet \cite{DBLP:conf/cvpr/HeZRS16} & \multirow{6}[0]{*}{ResNet-101} & 44.549M & 7.849 & 78.200  & 93.906 \\
				SENet \cite{DBLP:conf/cvpr/HuSS18} &       & 49.327M & 7.863 & 78.468 & 94.102 \\
				CBAM \cite{DBLP:conf/eccv/WooPLK18} &       & 49.330M & 7.879 & 78.354 & 94.064 \\
				SGE-Net \cite{DBLP:journals/corr/abs-1905-09646} &       & 44.553M & 7.858 & 78.798 & 94.368 \\
				ECA-Net \cite{wang2020eca} &       & \textbf{44.549M} & 7.858 & 78.650 & 94.340 \\
				\textbf{SA-Net (Ours)} &       & 44.551M & \textbf{7.854} & \textbf{78.960 (\textcolor[rgb]{ 0,  0,  1}{$\uparrow 0.76$})} & \textbf{94.492 (\textcolor[rgb]{ 0,  0,  1}{$\uparrow 0.59$})} \\
				\bottomrule[1.2pt]
			\end{tabular}
		}
	\end{center}
\end{table*}

{\bf Classification on ImageNet-1k.}
We compare SA-Net with the current SOTA attention methods. Evaluation metrics include both efficiency(i.e., network parameters and GFLOPs) and effectiveness(i.e., Top-1/Top-5 accuracy). As shown in Table \ref{tab:tab1}, SA-Net shares almost the same model complexity (i.e., network parameters and FLOPs) with the original ResNet~\cite{DBLP:conf/cvpr/HeZRS16}, but achieves 1.34\% gains in terms of Top-1 accuracy and 0.89\% advantages over Top-5 accuracy (on ResNet-50). When using ResNet-101 as the backbone, the performance gains are 0.76\% and 0.59\%, respectively. Comparing with SOTA counterparts, SA obtains higher accuracy while benefiting lower or similar model complexity. 
Specifically, when using ResNet-101 as backbone, the recently SOTA SE~\cite{DBLP:conf/cvpr/HuSS18} module increases 4.778M parameters, 14.34FLOPs, and 0.268 top-1 accuracy, but our SA only increase 0.002M parameters, 5.12FLOPs, and top-1 0.76 accuracy, which can demonstrate that SA is lighter and more efficient.

\begin{table}[h]
	\caption{Performance comparisons of SA-Net (using ResNet-50 as backbones) with four options (i.e., eliminating Group Norm, eliminating Channel Shuffle, eliminating $\mathcal{F}_c(\cdot)$ and utilizing $1\times 1$ Conv to replace $\mathcal{F}_c(\cdot)$) on ImageNet-1k in terms of GFLOPs and Top-1/Top-5 accuracy (in \%). The best records are marked in \textbf{bold}.
	}
	\label{tab:tab2}
	\begin{center}
		\setlength{\tabcolsep}{1.6mm}{
			\begin{tabular}{lccc}
				\toprule[1.2pt]
				Methods & GFLOPs & Top-1 Acc (\%) & Top-5 Acc (\%) \\
				\midrule[1.2pt]
				origin & 4.125 & \textbf{77.724} & 93.798 \\
				w/o\_gn & 4.125 & 77.372 & 93.804 \\
				w/o\_shuffle & 4.125 & 77.598 & 93.758 \\
				w/o\_$\mathcal{F}_c(\cdot)$ & 4.125 & 77.608 & \textbf{93.886} \\
				$1\times 1$ Conv & 4.140  & 77.684 & 93.840  \\
				\bottomrule[1.2pt]
			\end{tabular}
		}
	\end{center}
\end{table}

{\bf Ablation Study.} We report the ablation studies of SA-Net50 on ImageNet-1k, to thoroughly investigate the components of the SA. As shown in Table \ref{tab:tab2}, the performance drops significantly when eliminating Group Norm, which indicates the distribution of features generated by different samples from the same semantic group is inconsistent. It is difficult to learn robust significant coefficients without normalization. When eliminating channel shuffle, the performance drops a little, which demonstrates that information communication among different groups can enhance the features representation. There is no doubt if $\mathcal{F}_c(\cdot)$ is eliminated, the performance drops since $\mathcal{F}_c(\cdot)$ is adopted to enhance the representation of features. However, the performance does not improve if we replace $\mathcal{F}_c(\cdot)$ with $1\times 1$ Conv. This may because the number of channels in each sub-feature is too few, so that it is unnecessary to exchange information among different channels.

{\bf Object Detection on MS COCO.}
We further train the current SOTA detectors on the COCO train2017 split, and evaluate bounding box Average Precision (AP) for object detection, and mask AP for instance segmentation. We implement all detectors using the MMDetection toolkit with the default settings and trained them within 12 epochs(namely, `1× schedule’). For a fair comparison, we only replace the pre-trained backbones on ImageNet-1k and transfer them to MS COCO by fine-tuning, keeping the other components in the entire detector intact. As shown in Table \ref{tab:tab3}, integration of either SE block or the proposed SA module can improve the performance of object detection using either one-stage or two-stage detector by a clear margin. Meanwhile, our SA can outperform SE block with lower model complexity. Specifically, adopting Faster R-CNN~\cite{DBLP:conf/nips/RenHGS15} as the basic detector, SA outperforms SE by 1.0\% and 1.4\% in terms of AP by using ResNet-50 and ResNet-101, respectively. 
If we use RetinaNet as the base detector, the gains both increase by 1.5\%.

\begin{table*}[htb]
	\caption{Object detection results of different attention methods on COCO val2017. 
	The best records and the improvements are marked in \textbf{bold} and \textcolor[rgb]{ 0,  0,  1}{$\uparrow$}, respectively.
	}
	\label{tab:tab3}
	\begin{center}
		\setlength{\tabcolsep}{2.5mm}{
			\begin{tabular}{l|c|c|c|c|c|c|c|c|c}
				\toprule[1.2pt]
				Methods & Detectors & Param. & GFLOPs & AP50:95 & AP50  & AP75  & AP$_S$ & AP$_M$ & AP$_L$ \\
				\midrule[1.2pt]
				ResNet-50 & \multirow{6}[0]{*}{Faster R-CNN} & 41.53M & 207.07 & 36.4 & 58.4 & 39.1 & 21.5 & 40.0   & 46.6 \\
				+ SE  &       & 44.02M & 207.18 & 37.7 & 60.1 & 40.9 & 22.9 & 41.9 & 48.2 \\
				+ \textbf{SA (Ours)} &       & \textbf{41.53M} & \textbf{207.35} & \textbf{38.7 (\textcolor[rgb]{ 0,  0,  1}{$\uparrow 2.3$})} & \textbf{61.2} & \textbf{41.4} & \textbf{22.3} & \textbf{42.5} & \textbf{49.8} \\
				\cmidrule{1-1}\cmidrule{3-10}
				ResNet-101 &       & 60.52M & 283.14 & 38.5 & 60.3 & 41.6 & 22.3 & 43.0  & 49.8 \\
				+ SE  &       & 65.24M & 283.33 & 39.6 & 62.0  & 43.1 & 23.7 & 44.0  & 51.4 \\
				+ \textbf{SA (Ours)} &       & \textbf{60.53M} & \textbf{283.60} & \textbf{41.0(\textcolor[rgb]{ 0,  0,  1}{$\uparrow 2.5$})}  & \textbf{62.7} & \textbf{44.8} & \textbf{24.4} & \textbf{45.1} & \textbf{52.5} \\
				\midrule[1.2pt]
				
				ResNet-50 & \multirow{6}[0]{*}{Mask R-CNN} & 44.18M & 275.58 & 37.3 & 59.0  & 40.2 & 21.9 & 40.9 & 48.1 \\
				+ SE  &       & 46.67M & 275.69 & 38.7 & 60.9 & 42.1 & 23.4 & 42.7 & 50.0 \\
				+ \textbf{SA (Ours)} &       & \textbf{44.18M} & \textbf{275.86} & \textbf{39.4(\textcolor[rgb]{ 0,  0,  1}{$\uparrow 2.1$})} & \textbf{61.5} & \textbf{42.6} & \textbf{23.4} & \textbf{42.8} & \textbf{51.1} \\
				\cmidrule{1-1}\cmidrule{3-10}
				ResNet-101 &       & 63.17M & 351.65 & 39.4 & 60.9 & 43.3 & 23.0  & 43.7 & 51.4 \\
				+ SE  &       & 67.89M & 351.84 & 40.7 & 62.5 & 44.3 & 23.9 & 45.2 & 52.8 \\
				+ \textbf{SA (Ours)} &       & \textbf{63.17M} & \textbf{352.10} & \textbf{41.6(\textcolor[rgb]{ 0,  0,  1}{$\uparrow 2.2$})} & \textbf{63.0}  & \textbf{45.5} & \textbf{24.9} & \textbf{45.5} & \textbf{54.2} \\
				\midrule[1.2pt]
				
				ResNet-50 & \multirow{6}[0]{*}{RetinaNet} & 37.74M & 239.32 & 35.6 & 55.5 & 38.3 & 20.0   & 39.6 & 46.8 \\
				+ SE  &       & 40.25M & 239.43 & 36.0 & 56.7 & 38.3 & 20.5 & 39.7 & \textbf{47.7} \\
				+ \textbf{SA (Ours)} &       & \textbf{37.74M} & \textbf{239.60} & \textbf{37.5(\textcolor[rgb]{ 0,  0,  1}{$\uparrow 1.9$})} & \textbf{58.5} & \textbf{39.7} & \textbf{21.3} & \textbf{41.2} & 45.9 \\
				\cmidrule{1-1}\cmidrule{3-10}
				ResNet-101 &       & 56.74M & 315.39 & 37.7 & 57.5 & 40.4 & 21.1 & 42.2 & 49.5 \\
				+ SE  &       & 61.49M & 315.58 & 38.8 & 59.3 & 41.7 & 22.1 & 43.2 & 51.5 \\
				+ \textbf{SA (Ours)} &       & \textbf{56.64M} & \textbf{315.85} & \textbf{40.3(\textcolor[rgb]{ 0,  0,  1}{$\uparrow 2.6$})} & \textbf{61.2} & \textbf{43.2} & \textbf{23.2} & \textbf{44.4} & \textbf{53.5} \\
				\bottomrule[1.2pt]
		\end{tabular}}
	\end{center}
\end{table*}

\begin{table}[htb]
	\caption{Instance segmentation results of various state-of-the-arts attention modules using Mask R-CNN on COCO val2017. 
	}
	\label{tab:tab4}
	\setlength{\tabcolsep}{1.0mm}{
		\begin{center}
			\begin{tabular}{l|c|c|c|c|c|c}
				\toprule[1.2pt]
				Methods & AP50:95 & AP50  & AP75  & AP$_S$ & AP$_M$ & AP$_L$ \\
				\midrule[1.2pt]
				ResNet-50 & 34.2 & 55.9 & 36.2 & 18.2 & 37.5 & 46.3 \\
				+ SE  & 35.4 & 57.4 & 37.8 & 17.1 & 38.6 & \textbf{51.8} \\
				+ ECA & 35.6 & 58.1 & 37.7 & 17.6 & 39.0  & \textbf{51.8} \\
				+ SGE & 34.9 & 56.9 & 37.0 & 19.1 & 38.4 & 47.3 \\
				+ \textbf{SA (Ours)} & \textbf{36.1(\textcolor[rgb]{ 0,  0,  1}{$\uparrow 1.9$})} & \textbf{58.7} & \textbf{38.2} & \textbf{19.4} & \textbf{39.4} & 49.0 \\
				\midrule[1.2pt]
				
				ResNet-101 & 35.9 & 57.7 & 38.4 & 19.2 & 39.7 & 49.7 \\
				+ SE  & 36.8 & 59.3 & 39.2 & 17.2 & 40.3 & 53.6 \\
				+ ECA & 37.4 & 59.9 & 39.8 & 18.1 & 41.1 & \textbf{54.1} \\
				+ SGE & 36.9 & 59.3 & 39.4 & 20.0 & 40.8 & 50.1 \\
				+ \textbf{SA (Ours)} & \textbf{38.0(\textcolor[rgb]{ 0,  0,  1}{$\uparrow 2.1$})} & \textbf{60.0} & \textbf{40.3} & \textbf{20.8} & \textbf{41.2} & 51.7 \\
				\bottomrule[1.2pt]
			\end{tabular}
		\end{center}
	}
\end{table}

{\bf Instance Segmentation on MS COCO.}
Instance segmentation using Mask R-CNN~\cite{DBLP:conf/iccv/HeGDG17} on MS COCO are shown in Table \ref{tab:tab4}. 
As shown in Table \ref{tab:tab4}, SA module achieves clearly improvement over the original ResNet and performs better than other state-of-the-arts attention modules (i.e., SE block, ECA module, and SGE unit), with less model complexity. Particularly, SA module achieves more gains for small objects, which are usually more difficult to be correctly detected and segmented. These results verify our SA module has good generalization for various computer vision tasks.

\section{Conclusion}
In this paper, we propose a novel efficient attention module SA to enhance the representation power of CNN networks. SA first groups the dimensions of channel into multiple sub-features before processing them in parallel. Then, for each sub-feature, SA utilizes a Shuffle Unit to capture feature dependencies in both spatial and channel dimensions. Afterward, all sub-features are aggregated, and finally, we adopt a ``channel shuffle" operator to make information communication between different sub-features.
Experimental results demonstrate that our SA is an extremely lightweight plug-and-play block, which can significantly improve the performance of various deep CNN architectures.

In the future, we will further explore the spatial and channel attention modules of SA and adopt them into more CNN architectures, including the ShuffleNet family, the SKNet~\cite{DBLP:conf/cvpr/LiW0019}, and MobileNetV3~\cite{DBLP:conf/iccv/HowardPALSCWCTC19}.


\bibliographystyle{IEEEbib}
\bibliography{sanet}

\begin{thebibliography}{10}

\bibitem{DBLP:journals/corr/abs-1904-11492}
Yue Cao, Jiarui Xu, Stephen Lin, Fangyun Wei, and Han Hu,
\newblock ``Gcnet: Non-local networks meet squeeze-excitation networks and
  beyond,''
\newblock {\em CoRR}, vol. abs/1904.11492, 2019.

\bibitem{DBLP:conf/cvpr/LiW0019}
Xiang Li, Wenhai Wang, Xiaolin Hu, and Jian Yang,
\newblock ``Selective kernel networks,''
\newblock in {\em {IEEE} Conference on Computer Vision and Pattern Recognition,
  {CVPR} 2019, Long Beach, CA, USA, June 16-20, 2019}, 2019, pp. 510--519.

\bibitem{DBLP:conf/cvpr/FuLT0BFL19}
Jun Fu, Jing Liu, Haijie Tian, Yong Li, Yongjun Bao, Zhiwei Fang, and Hanqing
  Lu,
\newblock ``Dual attention network for scene segmentation,''
\newblock in {\em {IEEE} Conference on Computer Vision and Pattern Recognition,
  {CVPR} 2019, Long Beach, CA, USA, June 16-20, 2019}, 2019, pp. 3146--3154.

\bibitem{DBLP:journals/corr/abs-1809-00916}
Yuhui Yuan and Jingdong Wang,
\newblock ``Ocnet: Object context network for scene parsing,''
\newblock {\em CoRR}, vol. abs/1809.00916, 2018.

\bibitem{DBLP:journals/corr/abs-1903-10829}
HyunJae Lee, Hyo{-}Eun Kim, and Hyeonseob Nam,
\newblock ``{SRM} : {A} style-based recalibration module for convolutional
  neural networks,''
\newblock {\em CoRR}, vol. abs/1903.10829, 2019.

\bibitem{DBLP:conf/eccv/ZhaoZLSLLJ18}
Hengshuang Zhao, Yi~Zhang, Shu Liu, Jianping Shi, Chen~Change Loy, Dahua Lin,
  and Jiaya Jia,
\newblock ``Psanet: Point-wise spatial attention network for scene parsing,''
\newblock in {\em Computer Vision - {ECCV} 2018 - 15th European Conference,
  Munich, Germany, September 8-14, 2018, Proceedings, Part {IX}}, 2018, pp.
  270--286.

\bibitem{DBLP:journals/corr/abs-1907-13426}
Xia Li, Zhisheng Zhong, Jianlong Wu, Yibo Yang, Zhouchen Lin, and Hong Liu,
\newblock ``Expectation-maximization attention networks for semantic
  segmentation,''
\newblock {\em CoRR}, vol. abs/1907.13426, 2019.

\bibitem{DBLP:journals/corr/abs-1908-07678}
Zhen Zhu, Mengde Xu, Song Bai, Tengteng Huang, and Xiang Bai,
\newblock ``Asymmetric non-local neural networks for semantic segmentation,''
\newblock {\em CoRR}, vol. abs/1908.07678, 2019.

\bibitem{DBLP:journals/corr/abs-1811-11721}
Zilong Huang, Xinggang Wang, Lichao Huang, Chang Huang, Yunchao Wei, and Wenyu
  Liu,
\newblock ``Ccnet: Criss-cross attention for semantic segmentation,''
\newblock {\em CoRR}, vol. abs/1811.11721, 2018.

\bibitem{DBLP:conf/eccv/WooPLK18}
Sanghyun Woo, Jongchan Park, Joon{-}Young Lee, and In~So Kweon,
\newblock ``{CBAM:} convolutional block attention module,''
\newblock in {\em Computer Vision - {ECCV} 2018 - 15th European Conference,
  Munich, Germany, September 8-14, 2018, Proceedings, Part {VII}}, 2018, pp.
  3--19.

\bibitem{DBLP:conf/cvpr/WangWZLZH20}
Qilong Wang, Banggu Wu, Pengfei Zhu, Peihua Li, Wangmeng Zuo, and Qinghua Hu,
\newblock ``Eca-net: Efficient channel attention for deep convolutional neural
  networks,''
\newblock in {\em 2020 {IEEE/CVF} Conference on Computer Vision and Pattern
  Recognition, {CVPR} 2020, Seattle, WA, USA, June 13-19, 2020}. 2020, pp.
  11531--11539, {IEEE}.

\bibitem{DBLP:journals/corr/abs-1905-09646}
Xiang Li, Xiaolin Hu, and Jian Yang,
\newblock ``Spatial group-wise enhance: Improving semantic feature learning in
  convolutional networks,''
\newblock {\em CoRR}, vol. abs/1905.09646, 2019.

\bibitem{DBLP:conf/eccv/MaZZS18}
Ningning Ma, Xiangyu Zhang, Hai{-}Tao Zheng, and Jian Sun,
\newblock ``Shufflenet {V2:} practical guidelines for efficient {CNN}
  architecture design,''
\newblock in {\em Computer Vision - {ECCV} 2018 - 15th European Conference,
  Munich, Germany, September 8-14, 2018, Proceedings, Part {XIV}}, 2018, pp.
  122--138.

\bibitem{DBLP:conf/cvpr/ZhangZLS18}
Xiangyu Zhang, Xinyu Zhou, Mengxiao Lin, and Jian Sun,
\newblock ``Shufflenet: An extremely efficient convolutional neural network for
  mobile devices,''
\newblock in {\em 2018 {IEEE} Conference on Computer Vision and Pattern
  Recognition, {CVPR} 2018, Salt Lake City, UT, USA, June 18-22, 2018}. 2018,
  pp. 6848--6856, {IEEE} Computer Society.

\bibitem{DBLP:conf/cvpr/SzegedyVISW16}
Christian Szegedy, Vincent Vanhoucke, Sergey Ioffe, Jonathon Shlens, and
  Zbigniew Wojna,
\newblock ``Rethinking the inception architecture for computer vision,''
\newblock in {\em 2016 {IEEE} Conference on Computer Vision and Pattern
  Recognition, {CVPR} 2016, Las Vegas, NV, USA, June 27-30, 2016}, 2016, pp.
  2818--2826.

\bibitem{DBLP:conf/aaai/SzegedyIVA17}
Christian Szegedy, Sergey Ioffe, Vincent Vanhoucke, and Alexander~A. Alemi,
\newblock ``Inception-v4, inception-resnet and the impact of residual
  connections on learning,''
\newblock in {\em Proceedings of the Thirty-First {AAAI} Conference on
  Artificial Intelligence, February 4-9, 2017, San Francisco, California,
  {USA}}, 2017, pp. 4278--4284.

\bibitem{DBLP:conf/cvpr/HeZRS16}
Kaiming He, Xiangyu Zhang, Shaoqing Ren, and Jian Sun,
\newblock ``Deep residual learning for image recognition,''
\newblock in {\em 2016 {IEEE} Conference on Computer Vision and Pattern
  Recognition, {CVPR} 2016, Las Vegas, NV, USA, June 27-30, 2016}, 2016, pp.
  770--778.

\bibitem{DBLP:conf/nips/KrizhevskySH12}
Alex Krizhevsky, Ilya Sutskever, and Geoffrey~E. Hinton,
\newblock ``Imagenet classification with deep convolutional neural networks,''
\newblock in {\em Advances in Neural Information Processing Systems 25: 26th
  Annual Conference on Neural Information Processing Systems 2012. Proceedings
  of a meeting held December 3-6, 2012, Lake Tahoe, Nevada, United States},
  2012, pp. 1106--1114.

\bibitem{DBLP:conf/cvpr/SandlerHZZC18}
Mark Sandler, Andrew~G. Howard, Menglong Zhu, Andrey Zhmoginov, and
  Liang{-}Chieh Chen,
\newblock ``Mobilenetv2: Inverted residuals and linear bottlenecks,''
\newblock in {\em 2018 {IEEE} Conference on Computer Vision and Pattern
  Recognition, {CVPR} 2018, Salt Lake City, UT, USA, June 18-22, 2018}, 2018,
  pp. 4510--4520.

\bibitem{DBLP:conf/iccv/HowardPALSCWCTC19}
Andrew Howard, Ruoming Pang, Hartwig Adam, Quoc~V. Le, Mark Sandler, Bo~Chen,
  Weijun Wang, Liang{-}Chieh Chen, Mingxing Tan, Grace Chu, Vijay Vasudevan,
  and Yukun Zhu,
\newblock ``Searching for mobilenetv3,''
\newblock in {\em 2019 {IEEE/CVF} International Conference on Computer Vision,
  {ICCV} 2019, Seoul, Korea (South), October 27 - November 2, 2019}. 2019, pp.
  1314--1324, {IEEE}.

\bibitem{DBLP:conf/nips/SabourFH17}
Sara Sabour, Nicholas Frosst, and Geoffrey~E. Hinton,
\newblock ``Dynamic routing between capsules,''
\newblock in {\em Advances in Neural Information Processing Systems 30: Annual
  Conference on Neural Information Processing Systems 2017, 4-9 December 2017,
  Long Beach, CA, {USA}}, 2017, pp. 3856--3866.

\bibitem{DBLP:conf/iclr/HintonSF18}
Geoffrey~E. Hinton, Sara Sabour, and Nicholas Frosst,
\newblock ``Matrix capsules with {EM} routing,''
\newblock in {\em 6th International Conference on Learning Representations,
  {ICLR} 2018, Vancouver, BC, Canada, April 30 - May 3, 2018, Conference Track
  Proceedings}. 2018, OpenReview.net.

\bibitem{DBLP:conf/cvpr/HuSS18}
Jie Hu, Li~Shen, and Gang Sun,
\newblock ``Squeeze-and-excitation networks,''
\newblock in {\em 2018 {IEEE} Conference on Computer Vision and Pattern
  Recognition, {CVPR} 2018, Salt Lake City, UT, USA, June 18-22, 2018}, 2018,
  pp. 7132--7141.

\bibitem{DBLP:conf/cvpr/0004GGH18}
Xiaolong Wang, Ross~B. Girshick, Abhinav Gupta, and Kaiming He,
\newblock ``Non-local neural networks,''
\newblock in {\em 2018 {IEEE} Conference on Computer Vision and Pattern
  Recognition, {CVPR} 2018, Salt Lake City, UT, USA, June 18-22, 2018}, 2018,
  pp. 7794--7803.

\bibitem{DBLP:conf/eccv/WuH18}
Yuxin Wu and Kaiming He,
\newblock ``Group normalization,''
\newblock in {\em Computer Vision - {ECCV} 2018 - 15th European Conference,
  Munich, Germany, September 8-14, 2018, Proceedings, Part {XIII}}, 2018, pp.
  3--19.

\bibitem{DBLP:conf/iccv/SelvarajuCDVPB17}
Ramprasaath~R. Selvaraju, Michael Cogswell, Abhishek Das, Ramakrishna Vedantam,
  Devi Parikh, and Dhruv Batra,
\newblock ``Grad-cam: Visual explanations from deep networks via gradient-based
  localization,''
\newblock in {\em {IEEE} International Conference on Computer Vision, {ICCV}
  2017, Venice, Italy, October 22-29, 2017}. 2017, pp. 618--626, {IEEE}
  Computer Society.

\bibitem{DBLP:journals/corr/GoyalDGNWKTJH17}
Priya Goyal, Piotr Doll{\'{a}}r, Ross~B. Girshick, Pieter Noordhuis, Lukasz
  Wesolowski, Aapo Kyrola, Andrew Tulloch, Yangqing Jia, and Kaiming He,
\newblock ``Accurate, large minibatch {SGD:} training imagenet in 1 hour,''
\newblock {\em CoRR}, vol. abs/1706.02677, 2017.

\bibitem{DBLP:conf/nips/RenHGS15}
Shaoqing Ren, Kaiming He, Ross~B. Girshick, and Jian Sun,
\newblock ``Faster {R-CNN:} towards real-time object detection with region
  proposal networks,''
\newblock in {\em NIPS 2015, December 7-12, 2015, Montreal, Quebec, Canada},
  Corinna Cortes, Neil~D. Lawrence, Daniel~D. Lee, Masashi Sugiyama, and Roman
  Garnett, Eds., 2015, pp. 91--99.

\bibitem{DBLP:conf/iccv/HeGDG17}
Kaiming He, Georgia Gkioxari, Piotr Doll{\'{a}}r, and Ross~B. Girshick,
\newblock ``Mask {R-CNN},''
\newblock in {\em {IEEE} International Conference on Computer Vision, {ICCV}
  2017, Venice, Italy, October 22-29, 2017}, 2017, pp. 2980--2988.

\end{thebibliography}

\end{document}